\documentclass{article}

\newif\ifcameraready
\camerareadytrue

\ifcameraready
  \usepackage[dblblindworkshop,final]{neurips_2026}
\else
  \usepackage[dblblindworkshop]{neurips_2026}
\fi
\workshoptitle{Long-Context Foundation Models}

\usepackage[utf8]{inputenc}       %
\usepackage[T1]{fontenc}
\usepackage{amsmath}              %
\usepackage{amssymb}
\usepackage[hidelinks]{hyperref}  %
\usepackage{url}

\mathchardef\UrlBigBreakPenalty=100
\mathchardef\UrlBreakPenalty=5000
\makeatletter
\newcommand*{\doi}{doi: \begingroup\urlstyle{rm}\scriptspace\z@
  \expandafter\def\expandafter\verbatim@nolig@list\expandafter{\verbatim@nolig@list\do\f}%
  \Url}
\makeatother
\usepackage{booktabs}             %
\usepackage{multirow}             %
\usepackage{graphicx}             %
\usepackage[expansion=false]{microtype}  %
\usepackage{xcolor}

\usepackage{pgfplots}
\pgfplotsset{compat=1.18}
\usetikzlibrary{arrows.meta}

\ifcameraready\else
  \makeatletter
  \renewcommand{\@noticestring}{Submitted to the Third Workshop on Long-Context Foundation Models at NeurIPS 2026. Do not distribute.}
  \makeatother
\fi

\title{Post-Training Agents for Long-Document Reasoning}
\hypersetup{pdftitle={Post-Training Agents for Long-Document Reasoning},
            pdfauthor={Sushant Mehta, Logan Ritchie, Edwin Chen}}

\author{%
  Sushant Mehta\thanks{Correspondence to: \texttt{research@surgehq.ai}.} \quad Logan Ritchie \quad Edwin Chen \\
  Surge AI%
}

\newcommand{\MthreeOne}{19}
\newcommand{\MthreeTwo}{26}

\begin{document}

\maketitle

\begin{abstract}
Professional documents are a long-context workload that retrieval-style benchmarks do not capture. In a ten-domain enterprise collection, a task's evidence is spread across a source document of median length 27 pages (mean 48.8, maximum 873); a third of tasks rest mainly on charts, drawings, and maps; and 70.9\% require two or more reasoning hops.
We post-train \mbox{Kimi-K2.7}, an open-weight foundation model, at its full context window on 1,465 expert-authored tasks with atomic rubrics from that collection, training rank-32 adapters with one round of privileged-context on-policy self-distillation followed by two rounds of rubric-reward reinforcement learning.
We render each PDF as text for training and as extracted text plus selected page images at evaluation (training on raw PDFs would need more than 256K tokens of context). The smoothed RL training reward rises from 0.37 in the first tenth of the run to 0.67 in the last.
We report accuracy together with the tokens and turns the model spends.
On the 100-task GDP.pdf holdout, served in a 262,144-token window, strict pass (all rubric criteria met) more than doubles after the second round, from 11.0\% to 24.5\%, with gains across input lengths up to 240K tokens. Mean tokens generated stay flat (9,803 to 9,733), so the gain does not come from longer outputs. \mbox{GPT-5.6}~Sol's leaderboard strict pass is 30.7\% at a measured 3,422 tokens.
On 220 out-of-domain GDPval tasks, where the model runs as a long-horizon tool-using agent, mean rubric reward rises from 0.571 to 0.746 after the third round and zero-credit tasks fall from 54 to 10, while messages per task fall 18\%, from 70.9 to 58.2. In paired trajectories, the third round replaces a context-thrashing retry loop (roughly 110 messages, the same statute re-extracted sixteen times) with one up-front check of all sources, and verification extends from checking that files exist to checking what they contain.
We argue that long-context evaluation of agents should report accuracy jointly with tokens and turns, and that datasets of long documents should ship text and image renderings alongside each PDF.

\end{abstract}

\section{Introduction}
\label{sec:intro}

Much long-context evaluation is built around locating evidence: find the needle, or answer the question whose evidence sits at a controlled depth~\citep{hsieh2024ruler,liu2024lost,kuratov2024babilong}.
Professional documents pose the problem differently.
The evidence for a benefits question may be a footnote three pages after the table it qualifies. A wiring question requires choosing the right diagram by serial-number range before tracing a wire across a page image, and a productivity rollup depends on which of two case volumes a formula applies to.
The difficulty lies in holding scope, exclusions, visual encodings, and definitions steady across a long, mixed-modality document.

We post-train \mbox{Kimi-K2.7}~\citep{moonshot2026kimik27,kimi2025k2}, an open-weight model, at its full context window on 1,465 such document tasks with expert rubrics.
Alongside accuracy, we measure the tokens it generates on the in-domain GDP.pdf holdout~\citep{garre2026gdppdf}, whose 100 public tasks are single-turn and disjoint from the training tasks, and the turns and messages it uses when run as a long-horizon agent on 220 GDPval tasks that require tools and deliverables~\citep{patwardhan2025gdpval}.
A companion report on the same training run presents the benchmark results after the first two rounds and an analysis of tool use on GDPval~\citep{ritchie2026companion}. This paper adds the workload analysis, the training dynamics, the third round, the tokens and turns the trained policy spends, and trajectory changes across rounds.
We write $M_0$ for the base model and $M_1$, $M_2$, and $M_3$ for the checkpoints after training rounds 1--3 (R1--R3 in tables): distillation, reinforcement learning (RL), and continued RL.

Our contributions are:
(i) a characterization of professional-document reasoning as a long-context multimodal workload, with page, hop, and modality distributions over a ten-domain collection, input sizes in tokens, and frontier pass rates that show it is unsolved (Section~\ref{sec:workload});
(ii) the post-training recipe and its dynamics: privileged-context on-policy self-distillation followed by rubric-reward RL, the mechanics that keep the teacher's longer prompt from truncating the student's response, and the training reward over the RL run (Section~\ref{sec:fitting});
(iii) measurements showing that post-training more than doubles in-domain strict pass (11.0\% to 24.5\%) at constant token spend, with gains across input lengths up to 240K tokens, and that out-of-domain scores rise as messages fall (Section~\ref{sec:economics});
(iv) measured and qualitative evidence of what changes in the trajectories: writing the deliverable, budget-aware research, checks on the content of the deliverable, and scope held across long documents (Section~\ref{sec:trajectories});
and (v) recommendations for long-context evaluation and dataset release (Section~\ref{sec:implications}).

\section{Related work}
\label{sec:related}

Long-context benchmarks range from synthetic suites that control the position and density of evidence~\citep{hsieh2024ruler,kuratov2024babilong} to multitask suites over realistic long inputs~\citep{bai2024longbench,bai2025longbenchv2,yen2025helmet}. LongProc tests long-form procedural generation~\citep{ye2025longproc}, and positional effects are well documented~\citep{liu2024lost}.
Document-understanding benchmarks address visually rich pages~\citep{mathew2021docvqa} and long, multi-page documents~\citep{ma2024mmlongbench,deng2025longdocurl,saadfalcon2024pdftriage}, and GDP.pdf catalogs the failure modes of frontier models on real professional PDFs~\citep{garre2026gdppdf}.
Long-horizon agentic evaluation adds tool use and deliverables~\citep{patwardhan2025gdpval,xu2024agentcompany,panavas2026handbook}, and work on evaluating agents and models argues for reporting cost alongside accuracy~\citep{kapoor2025agents,erol2026costofpass}.
On the training side, our recipe combines self-distillation from privileged information~\citep{zhao2026opsd,penaloza2026privileged}, a form of on-policy distillation~\citep{agarwal2024onpolicy,lu2025onpolicy}, with criterion-level rewards~\citep{gunjal2025rubrics,viswanathan2025checklists,mehta2026complexif,bhattarai2026rubric} and low-rank adaptation~\citep{hu2022lora,schulman2025lora}. Concurrently, \citet{wang2026opdbeforerl} warm-start rubric-based RL with on-policy distillation from a larger teacher that reads the rubric; our first round instead uses the student checkpoint as its own teacher, conditioned on written feedback about an earlier attempt.
Reinforcement learning on long inputs has been applied to question answering scored against reference answers~\citep{wan2025qwenlongl1,wang2026loongrl} and to agents that search and read the pages of long documents over multiple turns~\citep{lin2026mmdocr1,yang2026alden}; we train on single-turn professional tasks graded by expert rubrics and evaluate the trained model both on single-turn tasks and as a tool-using agent.
Document-parsing vendors report that supplying a structured parse alongside the original PDF raises frontier-model accuracy on GDP.pdf~\citep{reducto2026accuracy,pulse2026gdppdf}.

\section{Professional documents as a long-context workload}
\label{sec:workload}

The training corpus is drawn from an expert-authored collection of prompts paired with real professional PDFs across ten domains.
Each task supplies the source PDF, a workflow-realistic question written by a domain expert from work they had recently performed, and an atomic rubric (mean about 11 criteria, maximum 30; Appendix~\ref{app:rubric})\@. Each criterion is a single auditable statement with subjectiveness and severity metadata, typed as \emph{primary intent} (what the prompt leads the user to expect) or \emph{dodged bullet} (the absence of a likely error that the prompt does not mention).

\paragraph{Length.}
The 2,485 tasks profiled in the collection's catalog draw on 121,158 pages (Table~\ref{tab:pages}), with a median of 27 pages per task and a mean of 48.8.
Engineering and finance documents average more than 60 pages, and the longest document, at 873 pages, is from construction.
The 1,465 training tasks alone draw on 73,609 pages.
Under our evaluation rendering, the inputs of the 100 holdout tasks range from 6K to 240K tokens, with a median of 87K~\citep{ritchie2026companion}.

\begin{table}[t]
\caption{Source-page profile, by domain, of the enterprise collection that supplies the 1,465 training tasks. Values come from the collection's catalog at the time of the study; its rows cover 2,485 of the 3,001 tasks it listed.}
\label{tab:pages}
\centering
\small
\begin{tabular}{lrrrrr}
\toprule
Domain & Tasks & Total pages & Median & Mean & Max \\
\midrule
Engineering & 241 & 15,437 & 37 & 64.1 & 407 \\
Finance/Investing & 295 & 18,706 & 39 & 63.4 & 384 \\
Construction & 264 & 14,315 & 26 & 54.2 & 873 \\
Legal & 223 & 11,960 & 30 & 53.6 & 382 \\
Manufacturing/Supply chains & 248 & 12,448 & 31 & 50.2 & 302 \\
STEM/Research & 247 & 11,300 & 25 & 45.7 & 498 \\
Insurance & 233 & 10,295 & 26 & 44.2 & 552 \\
HR & 227 & 9,371 & 30 & 41.3 & 336 \\
Real estate & 236 & 8,489 & 20 & 36.0 & 475 \\
Healthcare & 271 & 8,837 & 16 & 32.6 & 375 \\
\midrule
All & 2,485 & 121,158 & 27 & 48.8 & 873 \\
\bottomrule
\end{tabular}
\end{table}

\begin{table}[t]
\caption{Reasoning hops and dominant evidence modality in the same catalog (hop counts for 2,485 tasks, evidence for 2,470), and frontier pass rates across the collection's prompts at that time. Parentheses give a model's reasoning-effort setting.}
\label{tab:hops}
\centering
\footnotesize
\begin{tabular}{lrr@{\hspace{5mm}}lrr@{\hspace{5mm}}lrr}
\toprule
Hops & Tasks & Share & Evidence & Tasks & Share & Model & pass@1 & pass@3 \\
\midrule
1 & 722 & 29.1\% & Mostly text & 1,246 & 50.4\% & \mbox{GPT-5.6}~Sol~(Medium) & 38.5\% & 48.7\% \\
2 & 471 & 19.0\% & Mostly visual & 851 & 34.5\% & Claude Opus 4.8 & 26.4\% & 35.1\% \\
3 & 313 & 12.6\% & Roughly even & 373 & 15.1\% & Gemini 3.5 Flash & 25.2\% & 35.0\% \\
4--5 & 373 & 15.0\% & & & & Gemini 3.1 Pro (High) & 24.0\% & 33.9\% \\
6--9 & 299 & 12.0\% & & & & Muse Spark 1.1 & 23.9\% & 33.3\% \\
10+ & 307 & 12.4\% & & & & Kimi~K2.6 & 19.6\% & 28.2\% \\
\bottomrule
\end{tabular}
\end{table}

\paragraph{Dispersion.}
Table~\ref{tab:hops} shows that 70.9\% of tasks require two or more reasoning hops, and 12.4\% require ten or more.
In the tasks we reviewed, hops are dependent lookups (a serial number that selects a diagram, a legend that decodes a fill color, a footnote that qualifies a table row), so the evidence for a single answer is distributed across the document and must be composed in order.

\paragraph{Modality.}
A third of tasks (34.5\%) rest mainly on visual evidence (charts, floor plans, schematics, maps), and a further 15.1\% draw evenly on text and images.
The long context therefore mixes extractable text with images, and reducing those images to text loses part of the encoding the question depends on.

The workload is unsolved at the frontier.
Across the collection's prompts, the best pass@1 was 38.5\% (\mbox{GPT-5.6}~Sol at medium reasoning effort) and pass@3 was 48.7\%; Kimi~K2.6 reached 19.6\%.
On the 100-task GDP.pdf holdout, the highest entry on the public leaderboard when this study ran (July 2026) was \mbox{GPT-5.6}~Sol~(Max), at 30.7\% strict pass, the share of tasks with every rubric criterion satisfied~\citep{garre2026gdppdf}. The benchmark keeps only tasks that at least two frontier models failed at collection time.
The recurring failures cataloged for that benchmark include table misalignment, chart misreading, ignored footnotes and exclusions, and superseded evidence~\citep{garre2026gdppdf}. We read these as failures to hold scope across a long document.

\section{Post-training recipe and dynamics}
\label{sec:fitting}

\subsection{Rendering documents for the model}
\label{sec:normalization}

Self-hosted open-weight models have no document input: the serving stack accepts text and, at most, images, so the harness must render each PDF itself.
Hosted APIs accept PDFs but convert them in provider-specific ways, typically to an image of each page plus extracted text~\citep{openai2026fileinputs,anthropic2026pdf,google2026documents}, so a comparison on raw PDFs measures a model together with its provider's document pipeline.
For training, OCR and text extraction turn each PDF into a Markdown rendering in which tables, charts, and figures are transcribed by a separate model (\mbox{GPT-5.4}~Mini), and maps and floor plans become structured records; without this step, training would have needed a context window of more than 256K tokens.
Training runs at the model's full context window; 20 of the 1,465 tasks, whose documents exceed 300 pages, were filtered out during training because they did not fit.

In the main evaluation setup, the user message holds the question, a relevance guide derived from the question (likely relevant pages with excerpts), the extracted text of the complete document, and up to 16 selected page images; it is served in the model's 262,144-token window.
Page images account for most of the input tokens at evaluation.

\begin{figure}[t]
\centering
\pgfplotsset{compat=1.18}%
\begin{tikzpicture}
\begin{axis}[
  width=0.99\linewidth,
  height=4.1cm,
  grid=major,
  grid style={gray!20},
  label style={font=\small},
  tick label style={font=\footnotesize},
  ylabel={Training reward},
  xmin=0, xmax=100,
  ymin=0.28, ymax=0.79,
  xtick={0,20,40,60,80,100},
  xticklabel={\pgfmathprintnumber{\tick}\%},
  ytick={0.3,0.4,...,0.7},
  yticklabel style={/pgf/number format/fixed, /pgf/number format/precision=1},
]
\addplot[blue!85!black, thick] table[x=x, y=y] {rl_reward.dat};
\addplot[only marks, mark=*, mark size=1.6pt, orange!85!black] coordinates {(96.344,0.7421)};
\node[font=\footnotesize, anchor=east] at (axis cs:95,0.75) {peak 0.74};
\end{axis}
\end{tikzpicture}
\caption{RL training reward: the severity-weighted rubric reward on training prompts, as an exponential moving average over 32 steps, plotted against the share of the logged run completed.}
\label{fig:reward}
\end{figure}
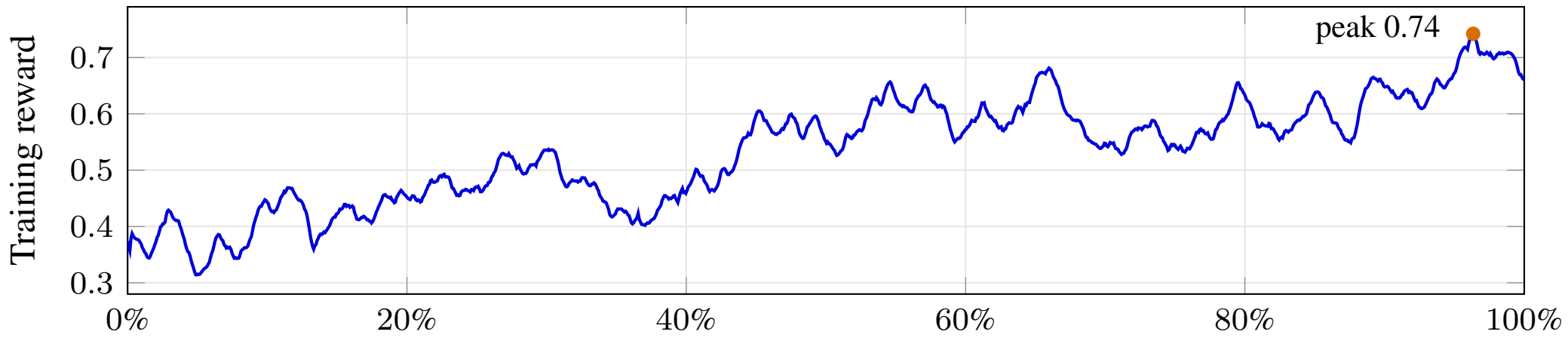

\subsection{Privileged-context distillation}
\label{sec:budgeting}

Round 1 of the recipe is privileged-context on-policy self-distillation.
Each round-1 task carries criterion-targeted feedback $z$ from \mbox{GPT-5.6}~Sol~(Max) on an earlier attempt, written before training; the student never sees $z$.
The student samples a response $y$ to the public task $x$; the same checkpoint, with the same active adapter, then acts as the teacher and scores the exact sampled tokens under the privileged prompt $(x, z)$. There is no task reward, so the token-level advantage is $-\lambda\,(\log \pi_\theta({y_t \mid x, y_{<t}}) - \mathrm{sg}[\log q_\theta({y_t \mid x, z, y_{<t}})])$, where $\lambda$ is a fixed coefficient, $q_\theta$ is the same network conditioned on the privileged prompt, and $\mathrm{sg}[\cdot]$ is a stop-gradient that treats the teacher's probabilities as fixed targets~\citep{zhao2026opsd,lu2025onpolicy}.

The teacher prompt is strictly longer than the student's, and the teacher must fit the whole student response after it. The student's maximum new tokens are therefore capped \emph{before} generation at the usable context limit minus the teacher prompt length, where the usable limit is the minimum of the rollout limit, the training sequence length, and the model's positional limit. A sample with no positive budget fails before generation rather than being truncated.
Aborted, empty, truncated, or over-context traces are rejected and never trained on as complete answers (the RL rounds likewise drop truncated samples and oversample with up to 16 retries). Teacher scoring is a prefill with zero new tokens over the teacher prompt and the student's recorded token IDs, so no decode-and-retokenize step can change what is scored.

\subsection{Rubric-reward reinforcement learning}
\label{sec:training}

Rounds 2 and 3 are rubric-reward RL: eight samples per prompt, with every rubric criterion judged by \mbox{GPT-5.4}~Mini at medium effort.
The reward (Figure~\ref{fig:reward}) is the severity-weighted fraction of criteria satisfied: primary-intent criteria weigh 1.0, 0.6, or 0.3 according to their severity label, dodged-bullet criteria enter as penalties rather than as credit, and the result is floored at zero (Appendix~\ref{app:config})\@.
The policy update uses group-relative advantages~\citep{shao2024deepseekmath,zheng2025gspo} with truncated importance sampling (TIS)~\citep{yao2025offpolicy} and asymmetric clipping (0.20/0.28)~\citep{yu2025dapo}.
Each response is capped at 65,536 tokens; the context is the model's full window.
Training runs on 64~H200 GPUs with context, tensor, pipeline, expert, and sequence parallelism and dynamic batching. It updates a rank-32 LoRA adapter over frozen base weights~\citep{hu2022lora}.
Round 1 uses 600 tasks and round 2 a disjoint 865, both balanced across the ten domains; round 3 continues RL from the round-2 adapter on the same 865-task pool under the same configuration (Table~\ref{tab:config}).

\subsection{Training dynamics}
\label{sec:dynamics}

Over the logged RL run (Figure~\ref{fig:reward}), the training reward rises from 0.37 at the start to a peak of 0.74 near the end and finishes at 0.66. Its mean is 0.37 over the first tenth of the run and 0.67 over the last.
The reward recovers to a new high after each of five falls of more than 0.10 below its running maximum, the largest from 0.68 to 0.53 about two-thirds of the way through.

\section{Results: accuracy, tokens, and turns}
\label{sec:economics}

\begin{table}[t]
\caption{In-domain holdout: the 100 public GDP.pdf tasks at pass@1, served in a 262,144-token window. The $M_2$ strict pass averages four evaluation runs (23, 24, 24, 27\%); the $M_2$ criterion pass rate comes from a single run, and all other cells are reported as single values. The \mbox{GPT-5.6}~Sol strict pass (public leaderboard, Max setting, July 2026) and criterion pass rate (from the leaderboard's evaluation data) were not produced under this harness; its token count is our measurement, so its passes-per-token ratio combines the two sources.}
\label{tab:indomain}
\centering
\footnotesize
\begin{tabular}{lcccc}
\toprule
Metric & Base $M_0$ & After R1 $M_1$ & After R2 $M_2$ & \mbox{GPT-5.6}~Sol \\
\midrule
Strict pass (all criteria) & 11.0\% & 13.0\% & 24.5\% & 30.7\% (leaderboard) \\
Criterion pass rate (per-task mean) & 0.6599 & 0.6786 & 0.7649 & 0.8260 \\
Tokens generated (mean) & 9,803 & 10,775 & 9,733 & 3,422 \\
Strict passes per 100K tokens & 1.12 & 1.21 & 2.52 & 8.97 \\
\bottomrule
\end{tabular}
\end{table}

\begin{table}[t]
\caption{Out-of-domain evaluation on 220 GDPval tasks (tools, web, deliverables), run through the Inspect Evals harness~\citep{inspect2024evals} on the Inspect AI framework~\citep{inspect2024ai} and judged by \mbox{GPT-5.4}~Mini. Reward is the share of rubric points earned; points missed are $(1-\mathrm{reward})$ times a rubric's net point total, with penalty items counted as negative. $M_3$ comes from the third of three evaluation runs (Section~\ref{sec:trajectories}).}
\label{tab:gdpval}
\centering
\footnotesize
\begin{tabular}{lccccc}
\toprule
Metric & Base $M_0$ & After R1 $M_1$ & After R2 $M_2$ & After R3 $M_3$ & \mbox{GPT-5.6}~Sol~(Medium) \\
\midrule
Perfect (all points) & 6 & 8 & 13 & 9 & 14 \\
$\leq$1 point missed & 21 & 23 & 32 & \MthreeOne & 30 \\
$\leq$2 points missed & 25 & 28 & 40 & \MthreeTwo & 39 \\
$\geq$90\% of points & 43 & 55 & 72 & 53 & 85 \\
Mean rubric reward & 0.571 & 0.742 & 0.703 & 0.746 & 0.761 \\
Zero-credit tasks & 54 & 10 & 29 & 10 & 14 \\
\bottomrule
\end{tabular}
\end{table}

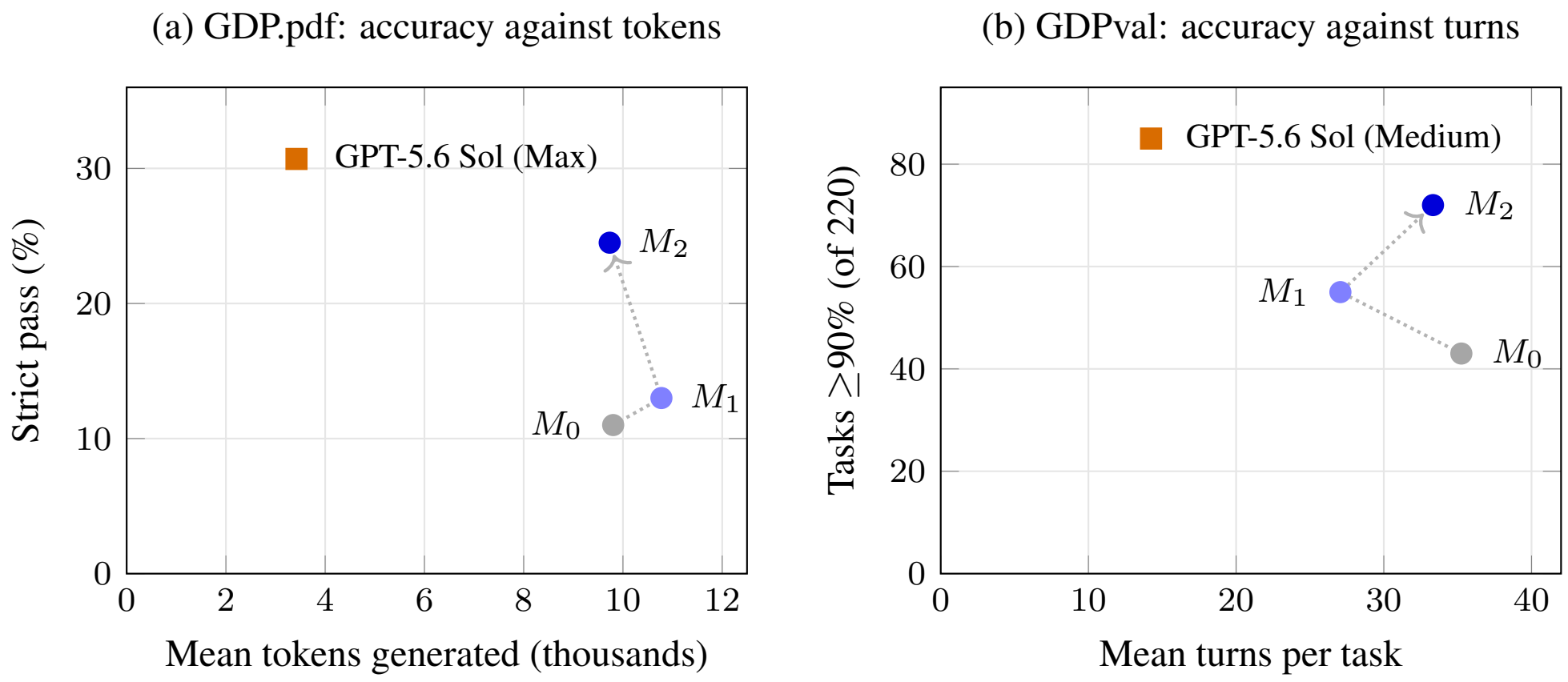
\begin{figure}[t]
\centering
\pgfplotsset{compat=1.18}%
\begin{tikzpicture}
\pgfplotsset{
  spendpanel/.style={
    width=0.48\linewidth,
    height=5.6cm,
    grid=major,
    grid style={gray!20},
    title style={font=\small},
    label style={font=\small},
    tick label style={font=\footnotesize},
  },
  spendpoint/.style={only marks, mark=*, mark size=2.4pt},
  spendpath/.style={gray!60, thick, densely dotted, ->, shorten >=3pt},
}
\tikzset{spendlabel/.style={font=\footnotesize}}
\begin{axis}[
  spendpanel,
  name=spendtokens,
  title={(a) GDP.pdf: accuracy against tokens},
  xlabel={Mean tokens generated (thousands)},
  ylabel={Strict pass (\%)},
  xmin=0, xmax=12.5,
  ymin=0, ymax=36,
  xtick={0,2,...,12},
  ytick={0,10,20,30},
]
\addplot[spendpath] coordinates {(9.803,11.0) (10.775,13.0) (9.733,24.5)};
\addplot[spendpoint, gray!70] coordinates {(9.803,11.0)};
\addplot[spendpoint, blue!50] coordinates {(10.775,13.0)};
\addplot[spendpoint, blue!85!black] coordinates {(9.733,24.5)};
\addplot[spendpoint, mark=square*, orange!85!black] coordinates {(3.422,30.7)};
\node[spendlabel, anchor=east] at (axis cs:9.5,11.0) {$M_0$};
\node[spendlabel, anchor=west] at (axis cs:11.05,13.0) {$M_1$};
\node[spendlabel, anchor=west] at (axis cs:10.0,24.5) {$M_2$};
\node[spendlabel, anchor=west] at (axis cs:3.9,30.7) {\mbox{GPT-5.6}~Sol~(Max)};
\end{axis}
\begin{axis}[
  spendpanel,
  at={(spendtokens.east)},
  anchor=west,
  xshift=1.6cm,
  title={(b) GDPval: accuracy against turns},
  xlabel={Mean turns per task},
  ylabel={Tasks $\geq$90\% (of 220)},
  xmin=0, xmax=42,
  ymin=0, ymax=95,
  xtick={0,10,...,40},
  ytick={0,20,...,80},
]
\addplot[spendpath] coordinates {(35.25,43) (27.06,55) (33.33,72)};
\addplot[spendpoint, gray!70] coordinates {(35.25,43)};
\addplot[spendpoint, blue!50] coordinates {(27.06,55)};
\addplot[spendpoint, blue!85!black] coordinates {(33.33,72)};
\addplot[spendpoint, mark=square*, orange!85!black] coordinates {(14.24,85)};
\node[spendlabel, anchor=west] at (axis cs:36.3,43) {$M_0$};
\node[spendlabel, anchor=east] at (axis cs:26.0,55) {$M_1$};
\node[spendlabel, anchor=west] at (axis cs:34.4,72) {$M_2$};
\node[spendlabel, anchor=west] at (axis cs:15.6,85) {\mbox{GPT-5.6}~Sol~(Medium)};
\end{axis}
\end{tikzpicture}
\caption{Accuracy against spend. (a) On the single-turn document interface, post-training moves the model upward at constant token spend; the frontier reference (leaderboard strict pass, measured output tokens) is more accurate and generates about a third as many tokens. (b) On the long-horizon agentic interface, turns (assistant messages) fall from 35.25 to 27.06 after the distillation round ($M_1$) and rise to 33.33, still below base, after the first RL round ($M_2$, the checkpoint with the most tasks at or above 90\% of rubric points); the reference completes tasks in fewer than half the turns.}
\label{fig:spend}
\end{figure}

\paragraph{In domain, accuracy per token more than doubles.}
On the 100-task holdout (Table~\ref{tab:indomain}), strict pass rises from 11.0\% at base to 13.0\% for $M_1$ and 24.5\% for $M_2$, and the criterion pass rate from 0.660 to 0.765 (a single run for $M_2$).
Mean tokens generated go from 9,803 to 10,775 and back down to 9,733, so $M_2$ earns more than twice as many strict passes per token as the base model at the same output length (2.52 versus 1.12 per hundred thousand tokens generated).
The strict-pass gain appears in all ten domains, and each of the four evaluation runs behind the 24.5\% figure (23\% to 27\%) exceeds the base run in a paired test (exact McNemar $p \leq 0.003$)~\citep{ritchie2026companion}.

\begin{table}[t]
\caption{Strict pass on the GDP.pdf holdout by evaluation input length in tokens~\citep{ritchie2026companion}; the $M_2$ figures are four-run means.}
\label{tab:length}
\centering
\footnotesize
\begin{tabular}{lrcc}
\toprule
Input tokens & Tasks & Base $M_0$ & After R2 $M_2$ \\
\midrule
Under 32K & 12 & 0.0\% & 12.5\% \\
32K--64K & 16 & 31.2\% & 46.9\% \\
64K--128K & 53 & 7.5\% & 18.4\% \\
Over 128K & 19 & 10.5\% & 30.3\% \\
\midrule
At or under 64K & 28 & 17.9\% & 32.1\% \\
Over 64K & 72 & 8.3\% & 21.5\% \\
\bottomrule
\end{tabular}
\end{table}

\paragraph{The gain holds across input lengths.}
Strict pass rises in every input-length bucket (Table~\ref{tab:length}), with the largest gain on the 19 inputs longer than 128K tokens (10.5\% to 30.3\%).
On the 72 inputs longer than 64K tokens it rises from 8.3\% to 21.5\%, and on the 28 shorter ones from 17.9\% to 32.1\%. The gain on the longer inputs minus the gain on the shorter ones is $-1.1$ points (95\% bootstrap interval $-13.0$ to $+9.7$), so there is no evidence that the gain shrinks with input length~\citep{ritchie2026companion}.

\paragraph{The frontier reference locates the efficiency gap.}
\mbox{GPT-5.6}~Sol's leaderboard strict pass is 30.7\% (Max setting) and its measured mean output is 3,422 tokens, or 8.97 strict passes per hundred thousand tokens if the two figures are combined (Figure~\ref{fig:spend}a).
The trained open model has closed much of the accuracy gap (24.5 versus 30.7) and almost none of the spend gap.
The open model still reasons at length over the rendered document where the reference answers briefly; whether that reflects how each model attends to the long input or how each has been trained to reason is a question this study cannot separate.

\paragraph{Out of domain, scores rise as messages fall.}
On GDPval, the policy runs a multi-turn tool loop that our post-training never exercised.
GDPval prompts are short (median 2.4K characters), so a session's long context comes from the model's own tool use~\citep{ritchie2026companion}.
A turn is one assistant message; messages also count tool results and the prompt, so they are about twice as many.
Mean rubric reward rises from 0.571 at base to 0.703 for $M_2$ and 0.746 for $M_3$, and zero-credit tasks fall from 54 to 29 and then 10 (Table~\ref{tab:gdpval}). Mean messages per task fall 18\%, from 70.9 at base to 58.2 for $M_3$ (67.0 for $M_2$).
Task by task, $M_2$ scores above base on 119 tasks and below it on 63 (Wilcoxon signed-rank $p < 10^{-6}$).
Its gain in the share of tasks reaching 90\% of rubric points extends to the four GDPval sectors with no counterpart among the training domains (20.0\% to 33.7\%) and to the 95 tasks with no input files (25\% to 40\%)~\citep{ritchie2026companion}.

\paragraph{Accuracy against turns.}
Tokens generated on GDPval do not fall as messages fall, so the trained models finish in fewer, denser turns.
After distillation, $M_1$ averages 27.06 turns against 35.25 at base and reaches the 90\% threshold on 55 tasks against 43; $M_2$ reaches it on 72 at 33.33 turns (Figure~\ref{fig:spend}b).
\mbox{GPT-5.6}~Sol~(Medium) reaches 85 tasks at the 90\% threshold in a mean of 14.24 turns, so the turn axis, like the token axis, has substantial headroom.

\paragraph{Different checkpoints lead on different metrics.}
From $M_2$ to $M_3$, tasks scoring more than 0.10 above base rise from 62 to 70 and tasks more than 0.10 below base fall from 24 to 18. The gains of $M_3$ over $M_2$ are concentrated at the bottom of the distribution (zero-credit tasks, 29 to 10) and in messages (67.0 to 58.2).
At the top, $M_2$ leads the trained checkpoints, with 13 perfect scores and 72 tasks at or above 90\%, against 9 and 53 for $M_3$.
After the distillation round alone, $M_1$ already reaches a mean reward of 0.742, with 10 zero-credit tasks and 70 tasks more than 0.10 above base.

\section{Analysis of the trajectories}
\label{sec:trajectories}

\begin{table}[t]
\caption{Session behavior on GDPval, base against $M_2$, in the companion report's replication of the GDPval comparison: a separate 100-task run, measured on the 94 task pairs in which both models wrote a gradable deliverable (60 pairs with named input files for the first row). Means per task unless stated. The paired 95\% bootstrap intervals of the changes in source coverage, read-only calls per task, retrieved text, unique share, and deliverable checks exclude zero~\citep{ritchie2026companion}.}
\label{tab:behavior}
\centering
\footnotesize
\begin{tabular}{lrr}
\toprule
Measure & Base $M_0$ & After R2 $M_2$ \\
\midrule
Named source files accessed before building & 76.1\% & 89.9\% \\
Read-only tool calls per write call & 2.77 & 1.72 \\
Environment text retrieved (tokens) & 13,459 & 7,172 \\
Share of retrieved text that is unique & 67.0\% & 73.9\% \\
Reasoning messages & 26.4 & 18.8 \\
Words per reasoning message & 214 & 486 \\
Tasks with an executable check of the deliverable & 5.3\% & 14.9\% \\
\bottomrule
\end{tabular}
\end{table}

In the companion report's GDPval replication (Table~\ref{tab:behavior}), $M_2$ covered more of the named sources than the base model while making fewer read-only tool calls, retrieved less text and repeated less of it, wrote fewer and longer reasoning messages, and checked its deliverable more often~\citep{ritchie2026companion}.
The paired trajectories below draw on seven GDPval tasks at $M_0$ and $M_3$, three of them also at $M_2$, and on five GDP.pdf tasks at $M_0$ and $M_2$ (scores are rubric fractions from single runs).

\paragraph{Writing the deliverable.}
Some of the reviewed base trajectories end with a complete, correct solution in a final reasoning message and never issue the tool call that would write the file.
On a leasing task, the base model's final message contained every resident, every date, and the planned table layout but ended in stray tool-call tokens with no call issued (0.000). $M_3$ built both deliverables in one scripted step and read its own PDFs back through a text extractor before finishing (1.000; $M_2$ also scored 1.000).
We set aside two of the three evaluation runs of $M_3$ because of a related failure (roughly 40\% of their trajectories ended after a single reasoning-only message) and report the third.

\paragraph{Covering the whole task.}
In two examples from the companion report, the base model treated one part as the whole. On a GDP.pdf parts question it took one of six seaming stations for the whole machine, and on a GDPval workload task it computed monthly totals from the first of four weekly timekeeping sheets. In both cases $M_2$ first established the full extent of the task~\citep{ritchie2026companion}.

\paragraph{Context thrashing on blocked sources.}
A legal-memo task required a statute, an amending bill, and case law. The base model never crossed from research to writing (0.000).
$M_2$ spent roughly 110 messages chasing sources that returned 403s and CAPTCHA walls, re-extracting the identical statutory text sixteen times, before it drafted and verified the memo (0.810).
$M_3$ checked every source in a single message before committing effort and applied an explicit fallback rule (the blocked cases were available in synopsis form, which sufficed for the one required holding). It drafted early, expanded the draft with the remaining budget, and verified the memo by reading it back (0.905).

\paragraph{Staged, restartable builds.}
On a CAD task (STEP solids, drawings, rendered PDF), $M_2$ built all geometry and then wrote a blind 523-line drawing script in one shot. When the script failed, the model got stuck inspecting its own source line numbers and ended with no drawings (0.000).
$M_3$ tested the SVG-to-PDF chain on a throwaway 10\,mm box before depending on it and asserted validity after nearly every boolean operation. After every patch it re-ran an idempotent build that regenerates the whole deliverable, so any crash cost only one iteration. It finished by re-importing the STEP files to check volumes, reading back the PDF text, and listing every output file (0.941).

\paragraph{Checking the deliverable.}
Across the replication, the share of tasks with an executable check of the written deliverable nearly triples (Table~\ref{tab:behavior}).
On a journal-club scheduling task, $M_3$ collected all nine articles through heavy rate limiting with escalating backoff, validated each download by its file signature and by text it expected to find, caught that one month's PDF was an unrelated 2022 report, and replaced it with the correct article before finishing.
On a residency-tracking workbook, $M_2$ chose a standard-deviation convention without checking it and flagged four residents where two were intended (0.776); $M_3$ computed the flags under both the sample and the population convention, compared them, and kept the one that matched the intended set (0.836).

\paragraph{Scope errors in long documents.}
On each of the five GDP.pdf pairs, $M_2$ corrected a scope error that the base model made (Appendix~\ref{app:pairs})\@. It distinguished the designs plotted on the requested summary graphs from the larger implementation tables elsewhere in the document, and kept a medication lookup within the named table rather than importing associations from a broader one. It also used a serial number to select the correct wiring diagram before tracing a wire across the page image, read a map by its legend's fill encoding rather than nearby prose, and defined the applicable case volume before applying a savings formula.

\section{Implications for long-context evaluation and datasets}
\label{sec:implications}

\paragraph{Report accuracy with spend.}
The trained model and the frontier reference differ most in spend, which accuracy alone does not show.
We recommend reporting tokens generated per task on single-turn interfaces, and turns and messages per task on agentic ones, alongside pass rates~\citep{kapoor2025agents}, and describing improvements as movements in the plane of accuracy against spend (Figure~\ref{fig:spend}).

\paragraph{Use thresholds and multiple runs.}
Strict pass on the 100-task holdout moved by four points across four runs of the same checkpoint (23--27\%).
Reported alongside a mean score, threshold metrics (perfect, at most one or two points missed, at least 90\%) show whether a gain comes from near-misses or from complete passes.

\paragraph{Use one document rendering when comparing models.}
Providers ingest PDFs differently, so comparisons on raw documents also compare document pipelines.

\paragraph{Ship text and image renderings with long-document datasets.}
We recommend that dataset builders release, alongside each PDF, a Markdown rendering of the text with references to separate image files for each figure, chart, or drawing.
Users who do not need the native PDF would train and iterate faster, and the visual evidence a third of tasks depend on would remain as images.

\paragraph{Guard long-prompt training against silent failures.}
Reserving the response budget against the longest prompt that must see the response keeps a response from being cut to fit a teacher. Rejecting truncated samples stops a truncated answer from being learned as complete, and teacher prefill over the exact token IDs avoids scoring a retokenized response under a shifted alignment.

\section{Conclusion}
\label{sec:conclusion}

Real professional documents are long, their evidence is dispersed, and they mix text with images. Frontier models do not yet solve them.
After two rounds of post-training at full context on 1,465 tasks over such documents, an open-weight model more than doubled its strict pass on the in-domain holdout at constant token spend, with gains across input lengths up to 240K tokens. On long-horizon agentic work that the post-training never exercised, it raised scores and cut zero-credit tasks from 54 to 10 after a third round, in 18\% fewer messages per task.
In the trajectories, the third round replaced a blind script with a restartable build, and a long retry loop with one up-front check of all sources and a fallback rule; verification extended to the deliverable's content, and after two rounds the model held a question's scope across pages and modalities.
The frontier reference, more accurate at roughly a third of the tokens and under half the turns, marks the remaining headroom on spend.

\bibliographystyle{plainnat}
\bibliography{references}

\begin{thebibliography}{51}
\providecommand{\natexlab}[1]{#1}
\providecommand{\url}[1]{\texttt{#1}}
\expandafter\ifx\csname urlstyle\endcsname\relax
  \providecommand{\doi}[1]{doi: #1}\else
  \providecommand{\doi}{doi: \begingroup \urlstyle{rm}\Url}\fi

\bibitem[Agarwal et~al.(2024)Agarwal, Vieillard, Zhou, Stanczyk, Ramos~Garea, Geist, and Bachem]{agarwal2024onpolicy}
Rishabh Agarwal, Nino Vieillard, Yongchao Zhou, Piotr Stanczyk, Sabela Ramos~Garea, Matthieu Geist, and Olivier Bachem.
\newblock On-policy distillation of language models: Learning from self-generated mistakes.
\newblock In \emph{The Twelfth International Conference on Learning Representations}, 2024.
\newblock URL \url{https://openreview.net/forum?id=3zKtaqxLhW}.

\bibitem[{Anthropic}(2026)]{anthropic2026pdf}
{Anthropic}.
\newblock {PDF} support.
\newblock Claude API documentation, 2026.
\newblock URL \url{https://platform.claude.com/docs/en/build-with-claude/pdf-support}.
\newblock Accessed November 2026.

\bibitem[Bai et~al.(2024)Bai, Lv, Zhang, Lyu, Tang, Huang, Du, Liu, Zeng, Hou, Dong, Tang, and Li]{bai2024longbench}
Yushi Bai, Xin Lv, Jiajie Zhang, Hongchang Lyu, Jiankai Tang, Zhidian Huang, Zhengxiao Du, Xiao Liu, Aohan Zeng, Lei Hou, Yuxiao Dong, Jie Tang, and Juanzi Li.
\newblock {LongBench}: A bilingual, multitask benchmark for long context understanding.
\newblock In \emph{Proceedings of the 62nd Annual Meeting of the Association for Computational Linguistics (Volume 1: Long Papers)}, pages 3119--3137. Association for Computational Linguistics, 2024.
\newblock \doi{10.18653/v1/2024.acl-long.172}.
\newblock URL \url{https://aclanthology.org/2024.acl-long.172/}.

\bibitem[Bai et~al.(2025)Bai, Tu, Zhang, Peng, Wang, Lv, Cao, Xu, Hou, Dong, Tang, and Li]{bai2025longbenchv2}
Yushi Bai, Shangqing Tu, Jiajie Zhang, Hao Peng, Xiaozhi Wang, Xin Lv, Shulin Cao, Jiazheng Xu, Lei Hou, Yuxiao Dong, Jie Tang, and Juanzi Li.
\newblock {LongBench} v2: Towards deeper understanding and reasoning on realistic long-context multitasks.
\newblock In \emph{Proceedings of the 63rd Annual Meeting of the Association for Computational Linguistics (Volume 1: Long Papers)}, pages 3639--3664. Association for Computational Linguistics, 2025.
\newblock \doi{10.18653/v1/2025.acl-long.183}.
\newblock URL \url{https://aclanthology.org/2025.acl-long.183/}.

\bibitem[Bhattarai et~al.(2026)Bhattarai, Boureima, Ranasinghe, Pakin, and O'Malley]{bhattarai2026rubric}
Manish Bhattarai, Ismael Boureima, Nishath~Rajiv Ranasinghe, Scott Pakin, and Dan O'Malley.
\newblock Rubric-grounded {RL}: Structured judge rewards for generalizable reasoning.
\newblock \emph{arXiv preprint arXiv:2605.08061}, 2026.
\newblock URL \url{https://arxiv.org/abs/2605.08061}.
\newblock AI for Science Workshop at ICML 2026.

\bibitem[Deng et~al.(2025)Deng, Yuan, Bu, Wang, Li, Xu, Li, Gao, Song, Zheng, and Liu]{deng2025longdocurl}
Chao Deng, Jiale Yuan, Pi~Bu, Peijie Wang, Zhong-Zhi Li, Jian Xu, Xiao-Hui Li, Yuan Gao, Jun Song, Bo~Zheng, and Cheng-Lin Liu.
\newblock {LongDocURL}: a comprehensive multimodal long document benchmark integrating understanding, reasoning, and locating.
\newblock In \emph{Proceedings of the 63rd Annual Meeting of the Association for Computational Linguistics (Volume 1: Long Papers)}, pages 1135--1159. Association for Computational Linguistics, 2025.
\newblock \doi{10.18653/v1/2025.acl-long.57}.
\newblock URL \url{https://aclanthology.org/2025.acl-long.57/}.

\bibitem[Erol et~al.(2026)Erol, El, Suzgun, Yuksekgonul, and Zou]{erol2026costofpass}
Mehmet~Hamza Erol, Batu El, Mirac Suzgun, Mert Yuksekgonul, and James Zou.
\newblock Cost-of-pass: An economic framework for evaluating language models.
\newblock In \emph{The Fourteenth International Conference on Learning Representations}, 2026.
\newblock URL \url{https://openreview.net/forum?id=vC9S20zsgN}.

\bibitem[Garre et~al.(2026)Garre, Ritchie, Mehta, and Chen]{garre2026gdppdf}
Suhaas Garre, Emily Ritchie, Sushant Mehta, and Edwin Chen.
\newblock {GDP.pdf}: Benchmarking grounded multimodal reasoning over professional {PDF} documents.
\newblock \emph{arXiv preprint arXiv:2607.11192}, 2026.
\newblock URL \url{https://arxiv.org/abs/2607.11192}.
\newblock Workshop on Knowledge-Intensive Multimodal Reasoning (KnowledgeMR) at CVPR 2026.

\bibitem[{Google}(2026)]{google2026documents}
{Google}.
\newblock Document understanding.
\newblock Gemini API documentation, 2026.
\newblock URL \url{https://aistudio.google.com/docs/document-processing}.
\newblock Accessed November 2026.

\bibitem[Gunjal et~al.(2026)Gunjal, Wang, Lau, Nath, He, Liu, and Hendryx]{gunjal2025rubrics}
Anisha Gunjal, Anthony Wang, Elaine Lau, Vaskar Nath, Yunzhong He, Bing Liu, and Sean Hendryx.
\newblock Rubrics as rewards: Reinforcement learning beyond verifiable domains.
\newblock In \emph{The Fourteenth International Conference on Learning Representations}, 2026.
\newblock URL \url{https://openreview.net/forum?id=c1bTcrDmt4}.

\bibitem[Hsieh et~al.(2024)Hsieh, Sun, Kriman, Acharya, Rekesh, Jia, Zhang, and Ginsburg]{hsieh2024ruler}
Cheng-Ping Hsieh, Simeng Sun, Samuel Kriman, Shantanu Acharya, Dima Rekesh, Fei Jia, Yang Zhang, and Boris Ginsburg.
\newblock {RULER}: What's the real context size of your long-context language models?
\newblock In \emph{First Conference on Language Modeling}, 2024.
\newblock URL \url{https://arxiv.org/abs/2404.06654}.

\bibitem[Hu et~al.(2022)Hu, Shen, Wallis, Allen-Zhu, Li, Wang, Wang, and Chen]{hu2022lora}
Edward~J. Hu, Yelong Shen, Phillip Wallis, Zeyuan Allen-Zhu, Yuanzhi Li, Shean Wang, Lu~Wang, and Weizhu Chen.
\newblock {LoRA}: Low-rank adaptation of large language models.
\newblock In \emph{The Tenth International Conference on Learning Representations}, 2022.
\newblock URL \url{https://openreview.net/forum?id=nZeVKeeFYf9}.

\bibitem[Kapoor et~al.(2025)Kapoor, Stroebl, Siegel, Nadgir, and Narayanan]{kapoor2025agents}
Sayash Kapoor, Benedikt Stroebl, Zachary~S. Siegel, Nitya Nadgir, and Arvind Narayanan.
\newblock {AI} agents that matter.
\newblock \emph{Transactions on Machine Learning Research}, 2025.
\newblock URL \url{https://openreview.net/forum?id=Zy4uFzMviZ}.

\bibitem[{Kimi Team}(2025)]{kimi2025k2}
{Kimi Team}.
\newblock {Kimi K2}: Open agentic intelligence.
\newblock \emph{arXiv preprint arXiv:2507.20534}, 2025.
\newblock URL \url{https://arxiv.org/abs/2507.20534}.

\bibitem[Kuratov et~al.(2024)Kuratov, Bulatov, Anokhin, Rodkin, Sorokin, Sorokin, and Burtsev]{kuratov2024babilong}
Yuri Kuratov, Aydar Bulatov, Petr Anokhin, Ivan Rodkin, Dmitry Sorokin, Artyom Sorokin, and Mikhail Burtsev.
\newblock {BABILong}: Testing the limits of {LLMs} with long context reasoning-in-a-haystack.
\newblock In \emph{Advances in Neural Information Processing Systems}, volume~37, pages 106519--106554, 2024.
\newblock \doi{10.52202/079017-3381}.
\newblock URL \url{https://proceedings.neurips.cc/paper_files/paper/2024/hash/c0d62e70dbc659cc9bd44cbcf1cb652f-Abstract-Datasets_and_Benchmarks_Track.html}.

\bibitem[Lin et~al.(2026)Lin, Hu, Wang, Zhou, Xi, Guo, Liu, Wang, Dou, Zhou, Yan, Han, Gui, Zhang, and Huang]{lin2026mmdocr1}
Jiahang Lin, Kai Hu, Binghai Wang, Yuhao Zhou, Zhiheng Xi, Honglin Guo, Shichun Liu, Junzhe Wang, Shihan Dou, Enyu Zhou, Hang Yan, Zhenhua Han, Tao Gui, Qi~Zhang, and Xuanjing Huang.
\newblock {MM-Doc-R1}: Training agents for long document visual question answering through multi-turn reinforcement learning.
\newblock In \emph{Findings of the Association for Computational Linguistics: {ACL} 2026}, pages 29770--29783. Association for Computational Linguistics, 2026.
\newblock \doi{10.18653/v1/2026.findings-acl.1488}.
\newblock URL \url{https://aclanthology.org/2026.findings-acl.1488/}.

\bibitem[Liu et~al.(2024)Liu, Lin, Hewitt, Paranjape, Bevilacqua, Petroni, and Liang]{liu2024lost}
Nelson~F. Liu, Kevin Lin, John Hewitt, Ashwin Paranjape, Michele Bevilacqua, Fabio Petroni, and Percy Liang.
\newblock Lost in the middle: How language models use long contexts.
\newblock \emph{Transactions of the Association for Computational Linguistics}, 12:\penalty0 157--173, 2024.
\newblock \doi{10.1162/tacl_a_00638}.
\newblock URL \url{https://aclanthology.org/2024.tacl-1.9/}.

\bibitem[Lu and {Thinking Machines Lab}(2025)]{lu2025onpolicy}
Kevin Lu and {Thinking Machines Lab}.
\newblock On-policy distillation.
\newblock \emph{Thinking Machines Lab: Connectionism}, 2025.
\newblock \doi{10.64434/tml.20251026}.
\newblock URL \url{https://thinkingmachines.ai/blog/on-policy-distillation/}.

\bibitem[Ma et~al.(2026)Ma, Zhang, Zhao, Song, Wang, Luo, and Sui]{ma2026routingreplay}
Wenhan Ma, Hailin Zhang, Liang Zhao, Yifan Song, Yudong Wang, Fuli Luo, and Zhifang Sui.
\newblock Stabilizing {MoE} reinforcement learning by aligning training and inference routers.
\newblock In \emph{Proceedings of the 43rd International Conference on Machine Learning}, volume 306 of \emph{Proceedings of Machine Learning Research}, pages 85320--85343. PMLR, 2026.
\newblock URL \url{https://proceedings.mlr.press/v306/ma26aw.html}.

\bibitem[Ma et~al.(2024)Ma, Zang, Chen, Chen, Jiao, Li, Lu, Liu, Ma, Dong, Zhang, Pan, Jiang, Wang, Cao, and Sun]{ma2024mmlongbench}
Yubo Ma, Yuhang Zang, Liangyu Chen, Meiqi Chen, Yizhu Jiao, Xinze Li, Xinyuan Lu, Ziyu Liu, Yan Ma, Xiaoyi Dong, Pan Zhang, Liangming Pan, Yu-Gang Jiang, Jiaqi Wang, Yixin Cao, and Aixin Sun.
\newblock {MMLongBench-Doc}: Benchmarking long-context document understanding with visualizations.
\newblock In \emph{Advances in Neural Information Processing Systems}, volume~37, pages 95963--96010, 2024.
\newblock \doi{10.52202/079017-3041}.
\newblock URL \url{https://proceedings.neurips.cc/paper_files/paper/2024/hash/ae0e43289bffea0c1fa34633fc608e92-Abstract-Datasets_and_Benchmarks_Track.html}.

\bibitem[Mathew et~al.(2021)Mathew, Karatzas, and Jawahar]{mathew2021docvqa}
Minesh Mathew, Dimosthenis Karatzas, and C.~V. Jawahar.
\newblock {DocVQA}: A dataset for {VQA} on document images.
\newblock In \emph{Proceedings of the IEEE/CVF Winter Conference on Applications of Computer Vision (WACV)}, pages 2200--2209, 2021.
\newblock \doi{10.1109/WACV48630.2021.00225}.
\newblock URL \url{https://openaccess.thecvf.com/content/WACV2021/html/Mathew_DocVQA_A_Dataset_for_VQA_on_Document_Images_WACV_2021_paper.html}.

\bibitem[Mehta et~al.(2026)Mehta, Panavas, Fleming, Mains, and Chen]{mehta2026complexif}
Sushant Mehta, Liudas Panavas, Eleanor Fleming, Paul Mains, and Edwin Chen.
\newblock {Complex-IF} and beyond: Expert rubrics for {RLVR}.
\newblock In \emph{Proceedings of the Fifth Workshop on Generation, Evaluation and Metrics ({GEM})}, pages 668--677. Association for Computational Linguistics, 2026.
\newblock \doi{10.18653/v1/2026.gem-main.61}.
\newblock URL \url{https://aclanthology.org/2026.gem-main.61/}.

\bibitem[{Moonshot AI}(2026)]{moonshot2026kimik27}
{Moonshot AI}.
\newblock {Kimi K2.7 Code}.
\newblock Hugging Face model card, 2026.
\newblock URL \url{https://huggingface.co/moonshotai/Kimi-K2.7-Code}.
\newblock Accessed November 2026.

\bibitem[{OpenAI}(2026)]{openai2026fileinputs}
{OpenAI}.
\newblock File inputs.
\newblock OpenAI API documentation, 2026.
\newblock URL \url{https://developers.openai.com/api/docs/guides/file-inputs}.
\newblock Accessed November 2026.

\bibitem[Panavas et~al.(2026)Panavas, Minus, Monton, Ray, Garre, Mehta, and Chen]{panavas2026handbook}
Liudas Panavas, Sebastian Minus, Bradley Monton, Derek Ray, Suhaas Garre, Sushant Mehta, and Edwin Chen.
\newblock {HANDBOOK.md}: A benchmark for long-context agentic instruction following.
\newblock \emph{arXiv preprint arXiv:2607.25398}, 2026.
\newblock URL \url{https://arxiv.org/abs/2607.25398}.
\newblock Workshop on Agent Behavior (WAB) at COLM 2026.

\bibitem[Patwardhan et~al.(2026)Patwardhan, Dias, Proehl, Kim, Wang, Watkins, Posada~Fishman, Aljubeh, Thacker, Fauconnet, Kim, Miserendino, Chabot, Li, Chao, Sharman, Barr, Glaese, and Tworek]{patwardhan2025gdpval}
Tejal Patwardhan, Rachel Dias, Elizabeth Proehl, Grace Kim, Michele Wang, Olivia Watkins, Sim{\'o}n Posada~Fishman, Marwan Aljubeh, Phoebe Thacker, Laurance Fauconnet, Natalie~S. Kim, Samuel Miserendino, Gildas Chabot, David Li, Patrick Chao, Michael Sharman, Alexandra Barr, Amelia Glaese, and Jerry Tworek.
\newblock {GDPval}: Evaluating {AI} model performance on real-world economically valuable tasks.
\newblock In \emph{The Fourteenth International Conference on Learning Representations}, 2026.
\newblock URL \url{https://openreview.net/forum?id=hcuEdq6eKD}.

\bibitem[Penaloza et~al.(2026)Penaloza, Vattikonda, Gontier, Lacoste, Charlin, and Caccia]{penaloza2026privileged}
Emiliano Penaloza, Dheeraj Vattikonda, Nicolas Gontier, Alexandre Lacoste, Laurent Charlin, and Massimo Caccia.
\newblock Privileged information distillation for language models.
\newblock In \emph{Proceedings of the 43rd International Conference on Machine Learning}, volume 306 of \emph{Proceedings of Machine Learning Research}, pages 97829--97868. PMLR, 2026.
\newblock URL \url{https://proceedings.mlr.press/v306/penaloza26a.html}.

\bibitem[{Pulse}(2026)]{pulse2026gdppdf}
{Pulse}.
\newblock {GDP.pdf} and the retrieval layer: How {Pulse} improves enterprise document {AI}.
\newblock Blog post, 2026.
\newblock URL \url{https://www.runpulse.com/blog/gdp-pdf-and-the-retrieval-layer}.
\newblock Accessed November 2026.

\bibitem[{RadixArk}(2026)]{radixark2026miles}
{RadixArk}.
\newblock {Miles} v0.1: Production-level post-training.
\newblock \emph{arXiv preprint arXiv:2609.08368}, 2026.
\newblock URL \url{https://arxiv.org/abs/2609.08368}.
\newblock Code: \url{https://github.com/radixark/miles}.

\bibitem[{Reducto}(2026)]{reducto2026accuracy}
{Reducto}.
\newblock {Reducto} raises frontier model accuracy on {GDP.pdf}.
\newblock Blog post, June 2026.
\newblock URL \url{https://reducto.ai/blog/reducto-raises-frontier-model-accuracy}.
\newblock Accessed November 2026.

\bibitem[Ritchie et~al.(2026)Ritchie, Mehta, Panavas, and Chen]{ritchie2026companion}
Logan Ritchie, Sushant Mehta, Liudas Panavas, and Edwin Chen.
\newblock Post-training {Kimi-K2.7} on {GDP.pdf}: Benchmark results and transfer to tool-using professional tasks on {GDPval}.
\newblock Technical report, Surge AI, September 2026.
\newblock URL \url{https://surgehq.ai/blog/gdp-pdf-post-training}.

\bibitem[Saad-Falcon et~al.(2024)Saad-Falcon, Barrow, Siu, Nenkova, Yoon, Rossi, and Dernoncourt]{saadfalcon2024pdftriage}
Jon Saad-Falcon, Joe Barrow, Alexa Siu, Ani Nenkova, Seunghyun Yoon, Ryan~A. Rossi, and Franck Dernoncourt.
\newblock {PDFTriage}: Question answering over long, structured documents.
\newblock In \emph{Proceedings of the 2024 Conference on Empirical Methods in Natural Language Processing: Industry Track}, pages 153--169. Association for Computational Linguistics, 2024.
\newblock \doi{10.18653/v1/2024.emnlp-industry.13}.
\newblock URL \url{https://aclanthology.org/2024.emnlp-industry.13/}.

\bibitem[Schulman and {Thinking Machines Lab}(2025)]{schulman2025lora}
John Schulman and {Thinking Machines Lab}.
\newblock {LoRA} without regret.
\newblock \emph{Thinking Machines Lab: Connectionism}, 2025.
\newblock \doi{10.64434/tml.20250929}.
\newblock URL \url{https://thinkingmachines.ai/blog/lora/}.

\bibitem[Shao et~al.(2024)Shao, Wang, Zhu, Xu, Song, Bi, Zhang, Zhang, Li, Wu, and Guo]{shao2024deepseekmath}
Zhihong Shao, Peiyi Wang, Qihao Zhu, Runxin Xu, Junxiao Song, Xiao Bi, Haowei Zhang, Mingchuan Zhang, Y.~K. Li, Y.~Wu, and Daya Guo.
\newblock {DeepSeekMath}: Pushing the limits of mathematical reasoning in open language models.
\newblock \emph{arXiv preprint arXiv:2402.03300}, 2024.
\newblock URL \url{https://arxiv.org/abs/2402.03300}.

\bibitem[Shoeybi et~al.(2019)Shoeybi, Patwary, Puri, LeGresley, Casper, and Catanzaro]{shoeybi2019megatron}
Mohammad Shoeybi, Mostofa Patwary, Raul Puri, Patrick LeGresley, Jared Casper, and Bryan Catanzaro.
\newblock {Megatron-LM}: Training multi-billion parameter language models using model parallelism.
\newblock \emph{arXiv preprint arXiv:1909.08053}, 2019.
\newblock URL \url{https://arxiv.org/abs/1909.08053}.

\bibitem[{UK AI Security Institute}(2024)]{inspect2024ai}
{UK AI Security Institute}.
\newblock {Inspect AI}: Framework for large language model evaluations.
\newblock GitHub repository, 2024.
\newblock URL \url{https://github.com/UKGovernmentBEIS/inspect_ai}.
\newblock Documentation at \url{https://inspect.aisi.org.uk/}. Accessed November 2026.

\bibitem[{UK AI Security Institute} et~al.(2024){UK AI Security Institute}, {Arcadia Impact}, and {Vector Institute}]{inspect2024evals}
{UK AI Security Institute}, {Arcadia Impact}, and {Vector Institute}.
\newblock {Inspect Evals}.
\newblock GitHub repository, 2024.
\newblock URL \url{https://github.com/UKGovernmentBEIS/inspect_evals}.
\newblock Maintained by Generality Labs. GDPval task: \url{https://ukgovernmentbeis.github.io/inspect_evals/evals/gdpval/}. Accessed November 2026.

\bibitem[Viswanathan et~al.(2025)Viswanathan, Sun, Ma, Kong, Cao, Neubig, and Wu]{viswanathan2025checklists}
Vijay Viswanathan, Yanchao Sun, Shuang Ma, Xiang Kong, Meng Cao, Graham Neubig, and Tongshuang Wu.
\newblock Checklists are better than reward models for aligning language models.
\newblock In \emph{Advances in Neural Information Processing Systems}, volume~38, 2025.
\newblock \doi{10.52202/085713-3824}.
\newblock URL \url{https://proceedings.neurips.cc/paper_files/paper/2025/hash/a6837c1dd021f76f1b4098e3722052a8-Abstract-Conference.html}.

\bibitem[Wan et~al.(2025)Wan, Shen, Liao, Shi, Li, Yang, Zhang, Huang, Zhou, and Yan]{wan2025qwenlongl1}
Fanqi Wan, Weizhou Shen, Shengyi Liao, Yingcheng Shi, Chenliang Li, Ziyi Yang, Ji~Zhang, Fei Huang, Jingren Zhou, and Ming Yan.
\newblock {QwenLong-L1}: Towards long-context large reasoning models with reinforcement learning.
\newblock \emph{arXiv preprint arXiv:2505.17667}, 2025.
\newblock URL \url{https://arxiv.org/abs/2505.17667}.

\bibitem[Wang et~al.(2026{\natexlab{a}})Wang, Zhang, Zhang, Shang, Yang, Chen, and Yang]{wang2026loongrl}
Siyuan Wang, Gaokai Zhang, Li~Lyna Zhang, Ning Shang, Fan Yang, Dongyao Chen, and Mao Yang.
\newblock {LoongRL}: Reinforcement learning for advanced reasoning over long contexts.
\newblock In \emph{The Fourteenth International Conference on Learning Representations}, 2026{\natexlab{a}}.
\newblock URL \url{https://openreview.net/forum?id=o29E01Q6bv}.

\bibitem[Wang et~al.(2026{\natexlab{b}})Wang, Shi, Chen, Zontak, He, and Pang]{wang2026opdbeforerl}
Xinpeng Wang, Wei Shi, Yu-Chia Chen, Maria Zontak, Yun He, and Richard~Yuanzhe Pang.
\newblock {OPD} before {RL}: Warm-starting rubric-based {RL} with on-policy distillation.
\newblock \emph{arXiv preprint arXiv:2610.02781}, 2026{\natexlab{b}}.
\newblock URL \url{https://arxiv.org/abs/2610.02781}.

\bibitem[Xu et~al.(2025)Xu, Song, Li, Tang, Jain, Bao, Wang, Zhou, Guo, Cao, Yang, Lu, Martin, Su, Maben, Mehta, Chi, Jang, Xie, Zhou, and Neubig]{xu2024agentcompany}
Frank~F. Xu, Yufan Song, Boxuan Li, Yuxuan Tang, Kritanjali Jain, Mengxue Bao, Zora~Z. Wang, Xuhui Zhou, Zhitong Guo, Murong Cao, Mingyang Yang, Hao~Yang Lu, Amaad Martin, Zhe Su, Leander~Melroy Maben, Raj Mehta, Wayne Chi, Lawrence Jang, Yiqing Xie, Shuyan Zhou, and Graham Neubig.
\newblock {TheAgentCompany}: Benchmarking {LLM} agents on consequential real world tasks.
\newblock In \emph{Advances in Neural Information Processing Systems}, volume~38, 2025.
\newblock \doi{10.52202/085713-0315}.
\newblock URL \url{https://proceedings.neurips.cc/paper_files/paper/2025/hash/0d744742f6fac4d1134c019b7cef3c8a-Abstract-Datasets_and_Benchmarks_Track.html}.

\bibitem[Yang et~al.(2026)Yang, Ruas, Tian, Wahle, Kurzawe, and Gipp]{yang2026alden}
Tianyu Yang, Terry Ruas, Yijun Tian, Jan~Philip Wahle, Daniel Kurzawe, and Bela Gipp.
\newblock {ALDEN}: Reinforcement learning for active navigation and evidence gathering in long documents.
\newblock In \emph{Proceedings of the 64th Annual Meeting of the Association for Computational Linguistics (Volume 1: Long Papers)}, pages 13371--13392. Association for Computational Linguistics, 2026.
\newblock \doi{10.18653/v1/2026.acl-long.611}.
\newblock URL \url{https://aclanthology.org/2026.acl-long.611/}.

\bibitem[Yao et~al.(2025)Yao, Liu, Zhang, Dong, Shang, and Gao]{yao2025offpolicy}
Feng Yao, Liyuan Liu, Dinghuai Zhang, Chengyu Dong, Jingbo Shang, and Jianfeng Gao.
\newblock Your efficient {RL} framework secretly brings you off-policy {RL} training.
\newblock Blog post, August 2025.
\newblock URL \url{https://fengyao.notion.site/off-policy-rl}.
\newblock Accessed November 2026.

\bibitem[Ye et~al.(2025)Ye, Yin, He, Zhang, Yen, Gao, Durrett, and Chen]{ye2025longproc}
Xi~Ye, Fangcong Yin, Yinghui He, Joie Zhang, Howard Yen, Tianyu Gao, Greg Durrett, and Danqi Chen.
\newblock {LongProc}: Benchmarking long-context language models on long procedural generation.
\newblock In \emph{Second Conference on Language Modeling}, 2025.
\newblock URL \url{https://openreview.net/forum?id=ruWC5LIMSo}.

\bibitem[Yen et~al.(2025)Yen, Gao, Hou, Ding, Fleischer, Izsak, Wasserblat, and Chen]{yen2025helmet}
Howard Yen, Tianyu Gao, Minmin Hou, Ke~Ding, Daniel Fleischer, Peter Izsak, Moshe Wasserblat, and Danqi Chen.
\newblock {HELMET}: How to evaluate long-context language models effectively and thoroughly.
\newblock In \emph{The Thirteenth International Conference on Learning Representations}, 2025.
\newblock URL \url{https://openreview.net/forum?id=293V3bJbmE}.

\bibitem[Yu et~al.(2025)Yu, Zhang, Zhu, Yuan, Zuo, Yue, Dai, Fan, Liu, Liu, Liu, Liu, Lin, Lin, Ma, Sheng, Tong, Zhang, Zhang, Zhang, Zhang, Zhu, Zhu, Chen, Chen, Wang, Yu, Song, Wei, Zhou, Liu, Ma, Zhang, Yan, Wu, and Wang]{yu2025dapo}
Qiying Yu, Zheng Zhang, Ruofei Zhu, Yufeng Yuan, Xiaochen Zuo, Yu~Yue, Weinan Dai, Tiantian Fan, Gaohong Liu, Juncai Liu, Lingjun Liu, Xin Liu, Haibin Lin, Zhiqi Lin, Bole Ma, Guangming Sheng, Yuxuan Tong, Chi Zhang, Mofan Zhang, Ru~Zhang, Wang Zhang, Hang Zhu, Jinhua Zhu, Jiaze Chen, Jiangjie Chen, Chengyi Wang, Hongli Yu, Yuxuan Song, Xiangpeng Wei, Hao Zhou, Jingjing Liu, Wei-Ying Ma, Ya-Qin Zhang, Lin Yan, Yonghui Wu, and Mingxuan Wang.
\newblock {DAPO}: An open-source {LLM} reinforcement learning system at scale.
\newblock In \emph{Advances in Neural Information Processing Systems}, volume~38, 2025.
\newblock \doi{10.52202/085713-3775}.
\newblock URL \url{https://proceedings.neurips.cc/paper_files/paper/2025/hash/a4277440d50f1f15d2cb4c14f7e0c0d2-Abstract-Conference.html}.

\bibitem[Zhao et~al.(2026)Zhao, Xie, Liu, Huang, Pang, Chen, and Grover]{zhao2026opsd}
Siyan Zhao, Zhihui Xie, Mengchen Liu, Jing Huang, Guan Pang, Feiyu Chen, and Aditya Grover.
\newblock Self-distilled reasoner: On-policy self-distillation for large language models.
\newblock In \emph{Proceedings of the 43rd International Conference on Machine Learning}, volume 306 of \emph{Proceedings of Machine Learning Research}, pages 162433--162448. PMLR, 2026.
\newblock URL \url{https://proceedings.mlr.press/v306/zhao26be.html}.

\bibitem[Zheng et~al.(2025)Zheng, Liu, Li, Chen, Yu, Gao, Dang, Liu, Men, Yang, Zhou, and Lin]{zheng2025gspo}
Chujie Zheng, Shixuan Liu, Mingze Li, Xiong-Hui Chen, Bowen Yu, Chang Gao, Kai Dang, Yuqiong Liu, Rui Men, An~Yang, Jingren Zhou, and Junyang Lin.
\newblock Group sequence policy optimization.
\newblock \emph{arXiv preprint arXiv:2507.18071}, 2025.
\newblock URL \url{https://arxiv.org/abs/2507.18071}.

\bibitem[Zheng et~al.(2024)Zheng, Yin, Xie, Sun, Huang, Yu, Cao, Kozyrakis, Stoica, Gonzalez, Barrett, and Sheng]{zheng2024sglang}
Lianmin Zheng, Liangsheng Yin, Zhiqiang Xie, Chuyue Sun, Jeff Huang, Cody~Hao Yu, Shiyi Cao, Christos Kozyrakis, Ion Stoica, Joseph~E. Gonzalez, Clark Barrett, and Ying Sheng.
\newblock {SGLang}: Efficient execution of structured language model programs.
\newblock In \emph{Advances in Neural Information Processing Systems}, volume~37, pages 62557--62583, 2024.
\newblock \doi{10.52202/079017-2000}.
\newblock URL \url{https://proceedings.neurips.cc/paper_files/paper/2024/hash/724be4472168f31ba1c9ac630f15dec8-Abstract-Conference.html}.

\bibitem[Zhu et~al.(2025)Zhu, Xie, Lv, and {slime Contributors}]{thudm2026slime}
Zilin Zhu, Chengxing Xie, Xin Lv, and {slime Contributors}.
\newblock {slime}: An {LLM} post-training framework for {RL} scaling.
\newblock GitHub repository, 2025.
\newblock URL \url{https://github.com/THUDM/slime}.
\newblock Accessed November 2026.

\end{thebibliography}

\appendix

\section{Training configuration}
\label{app:config}

\begin{table}[h]
  \caption{RL training configuration. The feedback model applies to the distillation round (R1), which also lowers the response cap to fit its privileged context.}
  \label{tab:config}
  \centering
  \scriptsize
  \begin{tabular}{llll}
    \toprule
    Field & Value & Field & Value \\
    \midrule
    Base model & \mbox{Kimi-K2.7} & Adapter & LoRA, rank 32, alpha 32, dropout 0 \\
    Adapter targets & \multicolumn{3}{l}{\texttt{q\_a\_proj}, \texttt{q\_b\_proj}, \texttt{kv\_a\_proj\_with\_mqa}, \texttt{kv\_b\_proj}, \texttt{o\_proj},} \\
     & \multicolumn{3}{l}{\texttt{gate\_proj}, \texttt{up\_proj}, \texttt{down\_proj}} \\
    Tasks & 600 (R1); 865 (R2, R3) & Batch size & 64 prompts per update \\
    Samples per prompt & 8 & Context & Full model window \\
    Response cap & 65,536 tokens & Max tokens per GPU & 32,768 (dynamic batching) \\
    Parallelism & TP8, PP4, CP2, EP16; sequence parallel & Temperature / top-$p$ & 0.7 / 0.95 \\
    Truncated samples & Dropped (oversampling retries: 16) & Reasoning effort & Maximum \\
    Criterion judge & \mbox{GPT-5.4}~Mini, medium effort & Feedback model (R1) & \mbox{GPT-5.6}~Sol~(Max) \\
    Advantage & Group-relative; TIS on & Clipping (low / high) & 0.20 / 0.28 \\
    Rollout routing replay & On & Entropy / KL coefficients & 0 / 0 \\
    Optimizer & Adam, $(\beta_1, \beta_2) = (0.9, 0.98)$ & Learning rate & $10^{-5}$, constant \\
    Weight decay & 0.1 & Gradient clipping & 0.1 \\
    Hardware & 8 nodes $\times$ 8 H200 & Checkpoint interval & Every 5 updates \\
    Backend & Megatron with FlashAttention & Seed & 42 \\
    \bottomrule
  \end{tabular}
\end{table}

The stack builds on slime for the distillation round, Miles for RL, SGLang for rollouts and teacher scoring, and Megatron for training~\citep{thudm2026slime,radixark2026miles,zheng2024sglang,shoeybi2019megatron}, with rollout routing replay for the mixture-of-experts routers~\citep{ma2026routingreplay}.
The teacher request supplies the teacher prompt's token IDs concatenated with the student's recorded response IDs, requests zero new tokens at temperature zero, and asks for log-probabilities starting one position before the response so that the first response token is included. The returned entries are checked against the response token IDs before use.
Private feedback is held in process-local runtime state that is excluded from serialization and cleared after scoring. The re-rendered public prompt must match the recorded prompt prefix token for token.
At evaluation, private context is stripped and teacher scoring does not run.

\paragraph{Model choice.}
We chose \mbox{Kimi-K2.7} as the strongest open-weight model on the in-domain benchmark at project start; a 30B-parameter alternative scored too low to provide a usable reward signal.

\paragraph{Reward.}
Each rubric criterion $i$ carries a weight $w_i$ set by its label: 1.0, 0.6, and 0.3 for primary-intent criteria marked as certain, possible, and unlikely dealbreakers, respectively, and 0.6 for dodged-bullet criteria.
The judge returns one binary verdict per criterion.
With $P$ the set of primary-intent criteria, $S \subseteq P$ those judged satisfied, and $D$ the dodged-bullet criteria to which the penalty applies, the reward is
\[
  r \;=\; \frac{\max\bigl(0,\; \sum_{i \in S} w_i \;-\; \sum_{i \in D} w_i\bigr)}{\sum_{i \in P} w_i},
\]
and $r = 0$ when a task has no primary-intent criteria.
Dodged-bullet criteria therefore never add credit and do not enter the denominator.

\section{Rubric size distribution}
\label{app:rubric}

In the catalog's rubric table (2,485 tasks), 584 tasks (23.5\%) have 1\nobreakdash--5 criteria, 928 (37.3\%) have 6\nobreakdash--10, 470 (18.9\%) have 11\nobreakdash--15, 235 (9.5\%) have 16\nobreakdash--20, 136 (5.5\%) have 21\nobreakdash--25, and 132 (5.3\%) have 26\nobreakdash--30.
Rubric size adds to a task's context demand as document length does, since more criteria mean more distinct facts to hold from more places in the document.

\section{Paired GDP.pdf comparisons}
\label{app:pairs}

We compare five base trajectories with $M_2$ trajectories on the same tasks, each from a single evaluation run chosen for the comparison; scores are rubric fractions.

\paragraph{Plotted population versus full tables (0.000 $\to$ 1.000).}
Asked for the highest-throughput designs under 1,500 LUTs on each of three summary graphs, excluding one named design, the base model drew on the full implementation tables elsewhere in the document and selected variants that were not the qualifying winners among the plotted designs.
The trained checkpoint treated each graph as a separate constrained maximization over the plotted population and returned the correct design, area, throughput, and variant feature for each.

\paragraph{Scope of a named table (0.333 $\to$ 1.000).}
Asked for medium-frequency animal models in a named table and the medications common to their associated disorders, the base model identified the models, then imported associations from a broader table and omitted two medications.
The trained checkpoint kept the lookup within the named table.

\paragraph{Diagram selection before tracing (0.500 $\to$ 1.000).}
Asked which replacement part connects to a fryer's I/O board through a specified wire, given a serial number, the base model inspected the wrong wiring context and reverse-matched a guess to the parts list.
The trained checkpoint used the serial range to select the correct diagram (manual page 2-35, PDF page 7), traced the wire from the I/O board connector to the transformer, and only then consulted the parts list.

\paragraph{Legend over prose (0.500 $\to$ 1.000).}
Asked which natural resource appears within a property boundary on an indicated map and how many instances there are, the base model saw two solid-blue features that the legend marked as lakes, then let nearby prose about wetlands and ponds override the map and added an unrequested acreage.
The trained checkpoint confined evidence to the indicated map and answered ``two lakes.''

\paragraph{Definition before formula (0.500 $\to$ 1.000).}
Asked for weekly mechanized-versus-manual productivity metrics, cost avoidance, payback, and projections, the base model applied the savings formula to all weekly cases rather than the mechanically processed cases, inflating savings and projections and shortening the payback estimate to seven months.
The trained checkpoint defined the applicable volume first, produced the corrected weekly figures and a payback of about 10.3 months, and showed its weekly rollups so that they could be checked.

\end{document}